\documentclass[10pt]{article} 

\usepackage[preprint]{rlj} 

\usepackage{amssymb}            
\usepackage{mathtools}          
\usepackage{mathrsfs}           
\usepackage{graphicx}           
\usepackage{subcaption}         
\usepackage[space]{grffile}     
\usepackage{url}                
\usepackage{lipsum}             
\usepackage{multirow}
\usepackage{multicol}
\usepackage{algorithm}    
\usepackage{algpseudocode} 
\newcommand{\showfont}{encoding: \f@encoding{},
  family: \f@family{},
  series: \f@series{},
  shape: \f@shape{},
  size: \f@size{}
}
\usepackage{amssymb}
\usepackage{amsfonts}
\usepackage{amsthm}
\usepackage{mathtools}
\usepackage{mathrsfs}
\usepackage{bm}
\usepackage{bbm}
\usepackage{algpseudocode}
\usepackage{tabularx}
\usepackage{hyperref}
\usepackage{cleveref}
\newcommand{\argmax}{\operatornamewithlimits{argmax}}

\theoremstyle{definition}

\newtheorem*{definition*}{Definision}
\numberwithin{definition}{subsection}
\crefname{definition}{Definition}{Definition}

\theoremstyle{definition}

\newtheorem*{proposition*}{Proposition}
\numberwithin{proposition}{subsection}
\crefname{proposition}{Proposition}{Proposition}

\theoremstyle{definition}

\newtheorem*{lemma*}{Lemma}
\numberwithin{lemma}{subsection}
\crefname{lemma}{Lemma}{Lemma}

\theoremstyle{definition}

\newtheorem*{corollary*}{Corollary}
\numberwithin{corollary}{subsection}
\crefname{corollary}{Corollary}{Corollary}

\newtheorem*{theorem*}{Theorem}
\numberwithin{theorem}{subsection}
\crefname{theorem}{Theorem}{Theorem}

\theoremstyle{definition}

\newtheorem*{assumption*}{Assumption}

\theoremstyle{remark}

\newtheorem*{example*}{\textbf{例}}
\numberwithin{example}{subsection}

\newcommand{\R}{\mathbb{R}} 

\newcommand{\E}[2][]{\mathbb{E}_{#1} \left[ {#2} \right]} 

\renewcommand{\epsilon}{\varepsilon}

\newcommand{\cA}{\mathcal{A}}

\newcommand{\cD}{\mathcal{D}}

\newcommand{\cM}{\mathcal{M}}

\newcommand{\cS}{\mathcal{S}}

\newcommand{\cX}{\mathcal{X}}

\title{Offline Reinforcement Learning with Domain-Unlabeled Data}

\setrunningtitle{Offline Reinforcement Learning with Domain-Unlabeled Data}

\author{Soichiro Nishimori\textsuperscript{1,2}, Xin-Qiang Cai\textsuperscript{2}, Johannes Ackermann\textsuperscript{1,2}, Masashi Sugiyama\textsuperscript{2,1}}

\emails{}

\affiliations{
$^{1}$\textbf{The University of Tokyo, Japan}\\
$^{2}$\textbf{RIKEN AIP, Japan}\\
}

\contribution{
We introduce Positive-Unlabeled Offline RL (PUORL), a novel offline RL setting with a small amount of data from a target domain and a large dataset containing data from multiple domains without domain labels. The goal is to learn a policy for the target domain.
}{
Existing cross-domain offline RL methods~\citep{liu2022dara, liuOODStateActions2023, wen2024contrastive} assume knowledge of the original domain of each transition, which is not accessible in our setting.
}

\contribution{
We propose a method that uses positive-unlabeled (PU) learning to filter the target-domain data from domain-unlabeled data.
}{
Our approach uses PU learning~\citep{li2003learning,kiryoPositiveUnlabeledLearningNonNegative2017} to classify domain-unlabeled samples as “positive” (target) or “negative” (other).
We then augment the labeled target-domain dataset with the domain-unlabeled samples predicted to be positive.
This filtering can be integrated with value-based offline RL algorithms.
}

\contribution{
We empirically demonstrate that our PU-based method accurately filters domain-unlabeled data and achieves high performance in a modified version of D4RL.
}{
We tested our approach on a modified D4RL benchmark~\citep{fu2021d4rl}, where only $1\%$--$3\%$ of samples contain domain labels, and the rest are domain-unlabeled, drawn from both the target and other domains with different dynamics.
Even with this limited labeling, our method closely matches an oracle baseline (which has access to all target-domain data) and overall achieves higher average returns than the other baselines, even under substantial dynamics mismatch.
}

\summary{
    Offline reinforcement learning (RL) is vital in areas where active data collection is expensive or infeasible, such as robotics or healthcare.
    In the real world, offline datasets often involve multiple “domains” that share the same state and action spaces but have distinct dynamics, and only a small fraction of samples are clearly labeled as belonging to the target domain we are interested in.
    For example, in robotics, precise system identification may only have been performed for part of the deployments.
    To address this challenge, we consider Positive-Unlabeled Offline RL (PUORL), a novel offline RL setting in which we have a small amount of labeled target-domain data and a large amount of domain-unlabeled data from multiple domains, including the target domain.
    For PUORL, we propose a plug-and-play approach that leverages positive-unlabeled (PU) learning to train a domain classifier. The classifier then extracts target-domain samples from the domain-unlabeled data, augmenting the scarce target-domain data.
    Empirical results on a modified version of the D4RL benchmark demonstrate the effectiveness of our method: even when only $1\%$--$3\%$ of the dataset is domain-labeled, our approach accurately identifies target-domain samples and achieves high performance, even under substantial dynamics shift.
    Our plug-and-play algorithm seamlessly integrates PU learning with existing offline RL pipelines, enabling effective multi-domain data utilization in scenarios where comprehensive domain labeling is prohibitive.
}
\begin{document}

\maketitle  

\begin{abstract}
    Offline reinforcement learning (RL) is vital in areas where active data collection is expensive or infeasible, such as robotics or healthcare.
    In the real world, offline datasets often involve multiple “domains” that share the same state and action spaces but have distinct dynamics, and only a small fraction of samples are clearly labeled as belonging to the target domain we are interested in.
    For example, in robotics, precise system identification may only have been performed for part of the deployments.
    To address this challenge, we consider Positive-Unlabeled Offline RL (PUORL), a novel offline RL setting in which we have a small amount of labeled target-domain data and a large amount of domain-unlabeled data from multiple domains, including the target domain.
    For PUORL, we propose a plug-and-play approach that leverages positive-unlabeled (PU) learning to train a domain classifier. The classifier then extracts target-domain samples from the domain-unlabeled data, augmenting the scarce target-domain data.
    Empirical results on a modified version of the D4RL benchmark demonstrate the effectiveness of our method: even when only $1\%$--$3\%$ of the dataset is domain-labeled, our approach accurately identifies target-domain samples and achieves high performance, even under substantial dynamics shift.
    Our plug-and-play algorithm seamlessly integrates PU learning with existing offline RL pipelines, enabling effective multi-domain data utilization in scenarios where comprehensive domain labeling is prohibitive.
\end{abstract}

\section{Introduction}
Offline reinforcement learning (RL)~\citep{Levine2020OfflineRL} trains policies exclusively from pre-collected datasets without further environmental interaction.
This paradigm has been applied to many real-world problems, including robotics~\citep{kalashnikovScalable, kalashnikov2021mtopt} and healthcare~\citep{Guez2008AdaptiveTO, killian2020empirical}, where live data collection is costly or infeasible.
This paper examines an offline RL setting where the dataset is collected in multiple \emph{domains}, environments that share the same state and action spaces but have different dynamics—with the goal of training a policy that performs well in a specific target domain.
In practice, however, annotating domain labels is labor-intensive or impractical at scale, resulting in a small amount of domain-labeled target data alongside a large volume of domain-unlabeled samples drawn from various domains, including the target domain.
One illustrative example arises in healthcare: if a specific disease significantly alters a patient’s response to treatment, it effectively changes the transition dynamics.
Only a small subset of patients are tested for disease with high cost of testing, leading to limited domain-labeled data and a predominance of domain-unlabeled samples~\citep{claesen2015building}.

Since offline RL depends on large, diverse datasets \citep{kalashnikov2021mtopt, padalkar2023open}, relying solely on the small domain-labeled subset may deteriorate policy performance.
Consequently, there is a pressing need to incorporate domain-unlabeled data effectively.
While recent studies have focused on enhancing target domain performance by utilizing data from a different domain~\citep{liu2022dara, wen2024contrastive, xuCrossDomainPolicyAdaptation2023}, these methods presuppose that clear domain labels are available for all samples, which does not hold in our setting.

To tackle this challenge, we propose a new offline RL setting called \textbf{P}ositive-\textbf{U}nlabeled \textbf{O}ffline \textbf{RL} (\textbf{PUORL}).
In PUORL, we have two types of data: a small amount of target-domain (positive-domain) data and a large volume of domain-unlabeled data, a mixture of samples from the positive domain and other domains (negative domains).
This setting is relevant in any setting where we aim to train agents based on a specific characteristic that significantly affects the dynamics.
This includes cases where a particular disease influences medical outcomes, as noted above, and scenarios such as unique road conditions in autonomous driving or a standard actuator defect in robotics~\citep{kiran2021deep,padakandla2021survey,shi2021offline}.

For PUORL, we propose a general approach that uses \emph{positive-unlabeled} (PU) learning~\citep{li2003learning, Bekker_2020,sugiyama2022machine} to train a classifier to distinguish positive-domain data from other domains (Sec.~\ref{sec: proposed method}).
Using the trained classifier, we filter out negative-domain data from a large, domain-unlabeled dataset, thereby augmenting the small domain-labeled data with additional positive-domain samples.
Then, we apply off-the-shelf offline RL algorithms to this augmented dataset.
Our framework functions as a plug-and-play module compatible with any value-based offline RL method, allowing users to adopt their preferred offline RL algorithm for PUORL.
Experiments on the modified version of the D4RL~\citep{fu2021d4rl}, where only $1\%$--$3\%$ of the data are domain-labeled, demonstrate that our method accurately identifies positive-domain data and effectively leverages the abundant domain-unlabeled dataset for offline RL (Sec.~\ref{sec: experiment}).
\paragraph{Related work.}
Cross-domain offline RL assumes fully domain-labeled datasets from two domains: a source domain with ample data and a target domain with fewer samples, where the goal is to effectively utilize the source domain data with different dynamics to improve the target domain performance~\citep{liu2022dara, xueStateRegularizedPolicy2023, xuCrossDomainPolicyAdaptation2023, wen2024contrastive, liuOODStateActions2023,lyu2024odrl}.
Some approaches fix the reward or filter transitions from a labeled source domain using discriminators~\citep{liu2022dara, wen2024contrastive}, while others constrain policies to remain in regions aligned with target-domain data~\citep{liuOODStateActions2023, xueStateRegularizedPolicy2023}.
Recently, \cite{lyu2024odrl} provided a benchmark for cross-domain offline RL.
In contrast to most methods, which assume available domain labels for all samples, our work handles a large amount of domain-unlabeled data, which may include samples from both target and non-target domains.
Please refer to App.~\ref{app: related_work} for the comprehensive related work.
\section{Preliminaries}
\label{sec:preliminaries}
\paragraph{Reinforcement learning (RL).}
RL \citep{suttonReinforcementLearningSecond2018} is characterized by a Markov decision process (MDP) \citep{putermanMarkovDecisionProcesses2014}, defined by 6-tuple: $\mathcal{M} := (\cS, \cA, P, p_0, R, \gamma)$.
Here, $\cS$ and $\cA$ denote the continuous state and action spaces, respectively.
$P: \cS \times \cA \times\cS\rightarrow [0, 1]$ defines the transition density, $p_0: \cS \rightarrow [0, 1]$ denotes the initial state distribution, $R: \cS\times\cA \rightarrow \R$ specifies the reward function, and $\gamma \in [0, 1)$ represents the discount factor.
In RL, the primary objective is to learn a policy $\pi: \cS \times \cA \rightarrow [0, 1]$, maximizing the expected cumulative discounted reward $\E[\pi, P]{\sum_{t=1}^{\infty} \gamma^t R(s_t, a_t)}$, where $\E[\pi, P]{\cdot}$ denotes the expectation over the sequence of states and actions $(s_1, a_1, \dots)$ generated by the policy $\pi$ and the transition density $P$.

In this paper, we assume that different domains correspond to distinct MDPs that differ only in their transition dynamics.
For example, two domains, $\mathcal{M}_1$ and $\mathcal{M}_2$, have different transition dynamics ($P_1$ and $P_2$), with the other components being the same.

\paragraph{Offline RL.}
To address the limitations on direct agent-environment interactions, offline RL \citep{Levine2020OfflineRL} employs a fixed dataset, $\cD := \{(s_i, a_i, r_i, s'_i)\}_{i=1}^{N}$, collected by a behavioral policy $\pi_\beta: \cS\times\cA \rightarrow [0, 1]$.
Let $\mu_\beta(s, a)$ be the stationary distribution over the state-action pair induced by the behavioral policy $\pi_\beta$.
The dataset $\cD$ is assumed to be generated as follows: $(s_i, a_i) \sim \mu_\beta(s, a)$, $r_i  = R(s_i, a_i)$, and $s'_i \sim P(\cdot|s_i, a_i)$.

\paragraph{Positive-unlabeled (PU) learning.}
PU learning is a method that trains a binary classifier using positive and unlabeled data \citep{li2003learning, bekkerEstimatingClassPrior2018, sugiyama2022machine}.
Let $X\in \R^d$ and $Y\in \{-1, +1\}$ be the random variables of the input and label in a binary classification problem.
We denote the data-generating joint density over $(X, Y)$ by $p(x, y)$. Let $p_\mathrm{p}(x):= p(x|Y=+1)$ and $p_\mathrm{n}(x):= p(x|Y=-1)$ be the densities of $x$ conditioned on the positive and negative labels respectively and $p(x):= \alpha_{\mathrm{p}} p_\mathrm{p}(x) + \alpha_{\mathrm{n}} p_\mathrm{n}(x)$ be the marginal density of the unlabeled data.
$\alpha_{\mathrm{p}}:= p(Y=+1)$ denotes the class prior probability (mixture proportion) for the positive label and $\alpha_{\mathrm{n}}:= p(Y=-1)= 1-\alpha_{\mathrm{p}}$ for the negative label.
In PU learning, we assume that we have two types of data: Positively labeled data $\cX_\mathrm{p}:= \{x^\mathrm{p}_i\}_{i=1}^{n_\mathrm{p}} \stackrel{\mathrm{i.i.d.}}{\sim}
 p_\mathrm{p}(x)$ and unlabeled data $\cX_\mathrm{u}:= \{x^\mathrm{u}_i\}_{i=1}^{n_\mathrm{u}} \stackrel{\mathrm{i.i.d.}}{\sim}
 p(x)$. The task of PU learning is to train a binary classifier $f :X \rightarrow \{-1, +1\}$ from positive data $\cX_\mathrm{p}$ and unlabeled data $\cX_\mathrm{u}$. Generally, PU learning methods require information on the mixture proportion ($\alpha_p$), and there are a bunch of mixture proportion estimation (MPE) methods \citep{IEICE:duPlessis+Sugiyama:2014,  scottRateConvergenceMixture2015,
 ML:duPlessis+etal:2017, gargMixtureProportionEstimation2021}.
Among the methods of PU learning, certain approaches, notably nnPU \citep{kiryoPositiveUnlabeledLearningNonNegative2017} and $(\mathrm{TED})^\mathrm{n}$ \citep{gargMixtureProportionEstimation2021}, demonstrate particular compatibility with neural networks.

\section{Method}
This section introduces a novel offline RL problem setting for leveraging domain-unlabeled data. We then propose a simple algorithm using PU learning to address this problem.
\subsection{Problem Formulation}
\label{sec:problem formulation}
\begin{figure}
    \centering
    \includegraphics[scale=0.27]{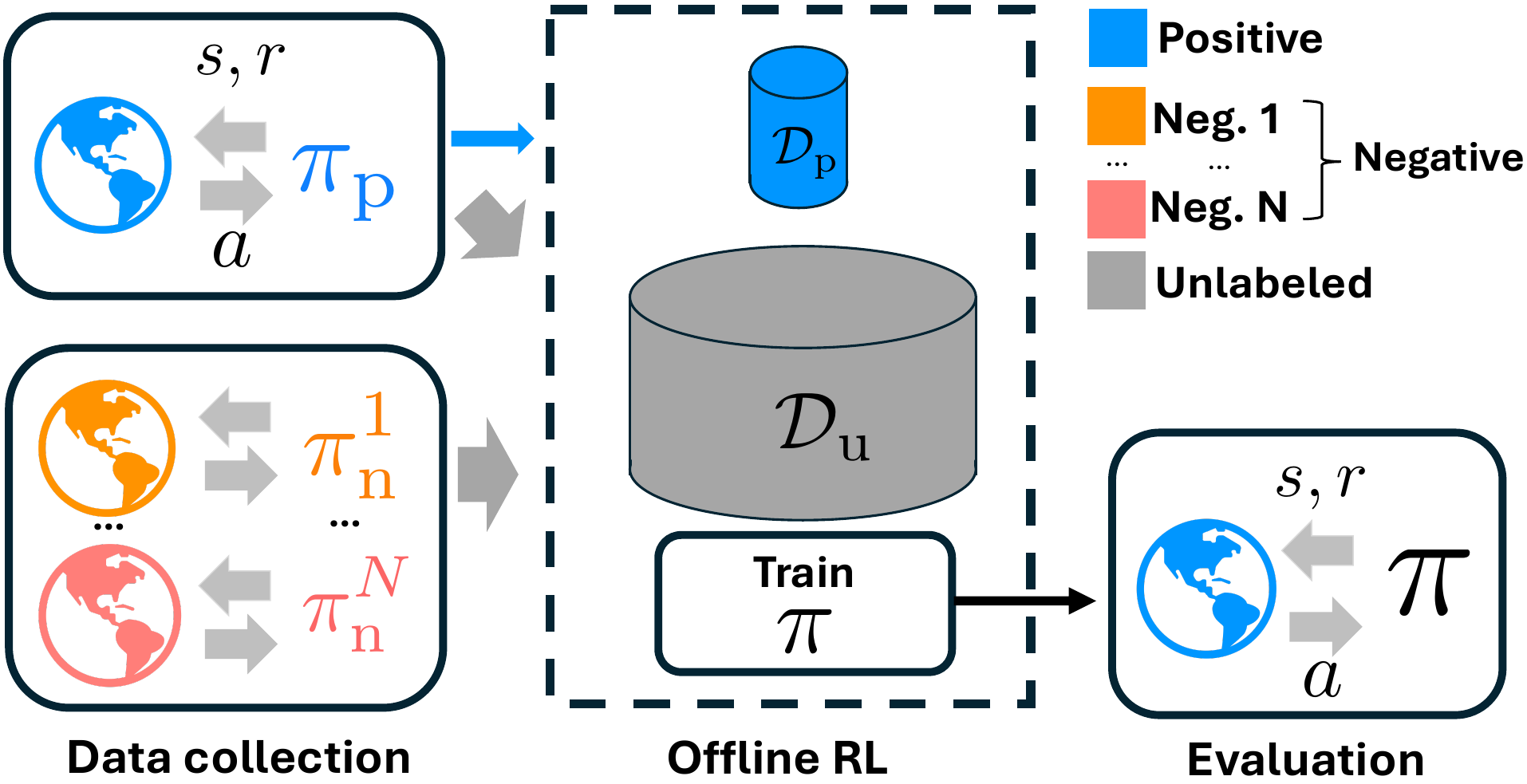}
    
    \caption{
 Diagram of Positive-Unlabeled Offline RL (PUORL). PUORL has a \textbf{positive domain} we target and \textbf{negative domains}, with different dynamics to the positive domain.
 We have two data types: \textbf{positive data} and \textbf{domain-unlabeled data}, which are mixtures of samples from the positive and negative domains.
 We train a policy to maximize the expected return in the positive domain.
 }
    \label{fig:formulation}
\end{figure}
We introduce \textbf{Positive-Unlabeled Offline RL (PUORL)} where the dataset is generated within multiple domains, with a small amount of data from one domain of our interest labeled and the rest provided as domain-unlabeled (Figure \ref{fig:formulation}).
In PUORL, we have a positive domain $\mathcal{M}_\mathrm{p}:= (\cS, \cA, P_{\mathrm{p}}, \rho, R, \gamma)$, for which we aim to maximize the expected return and negative domains $\{\mathcal{M}^{k}_\mathrm{n}:= (\cS, \cA, P^{k}_{\mathrm{n}}, \rho, R, \gamma)\}_{k=1}^{N}$, which share the same state and action spaces, initial state distribution, reward function and discount factor.
For each domain, there exist fixed behavioral policies: $\pi_{\mathrm{p}}$ for positive domain and $\pi^{k}_{\mathrm{n}}$ for negative domains, and they induce the stationary distributions over the state-action pair denoted as $\mu_\mathrm{p}(s, a)$ and $\mu^{k}_\mathrm{n}(s, a)$ for all $k \in \{1,\ldots, N\}$.
We define $\mu_\mathrm{n}(s, a) := \sum_{k=1}^{N} \eta_k \mu^{k}_\mathrm{n}(s, a)$, where $\eta_k \in [0, 1], \sum_{k=1}^{N} \eta_k = 1$ is the domain-mixture proportion.

We are given two datasets:
\begin{itemize}
    \item \textbf{Positive data}: explicitly labeled target-domain transitions, $\cD_\mathrm{p} := \{(s_i, a_i, r_i, s'_i)\}_{i=1}^{n_\mathrm{p}}$.
 These transitions are i.i.d. samples from $\mu_\mathrm{p}(s, a)$, $R$, and $P_\mathrm{p}$.

    \item \textbf{Domain-unlabeled data}: a mixture of positive and negative-domain transitions, $\cD_\mathrm{u} := \{(s_i, a_i, r_i, s'_i)\}_{i=1}^{n_\mathrm{u}}$.
 These transitions are i.i.d.~samples from $\mu_\mathrm{u}(s, a) := \alpha_\mathrm{p} \mu_\mathrm{p}(s, a) + \alpha_\mathrm{n} \mu_\mathrm{n}(s, a)$, $R$, and corresponding transition densites. We assume that $n_\mathrm{u} \gg n_\mathrm{p}$.
\end{itemize}
Henceforth, domain-unlabeled data will be referred to as \emph{unlabeled data} when it is clear from the context.
Although PUORL focuses on the difference in dynamics, we can generalize the problem set to encompass variations in the reward function.
Refer to Appendix~\ref{app: extension_to_reward_shift} for details.
Here, the objective is to learn the optimal policy in the positive domain of our interest as
\begin{align}
\label{single-domain:formulation}
    \pi^*(a|s) := \argmax_\pi \E[\pi, P_\mathrm{p}]{\sum_{t=1}^\infty \gamma^t R(s_t, a_t)}.
\end{align}
The most naive approach in this setup involves applying conventional offline RL methods on only a small amount of positive data $\cD_\mathrm{p}$.
However, using a small dataset increases the risk of encountering out-of-distribution state-action pairs due to the limited coverage of the dataset \citep{Levine2020OfflineRL}.
Conversely, utilizing all available data $\cD_\mathrm{p} \cup \cD_{\mathrm{u}}$ to increase the dataset size can hinder the agent's performance due to the different dynamics \citep{liu2022dara}.

\begin{figure}
    \centering
    \includegraphics[scale=0.3]{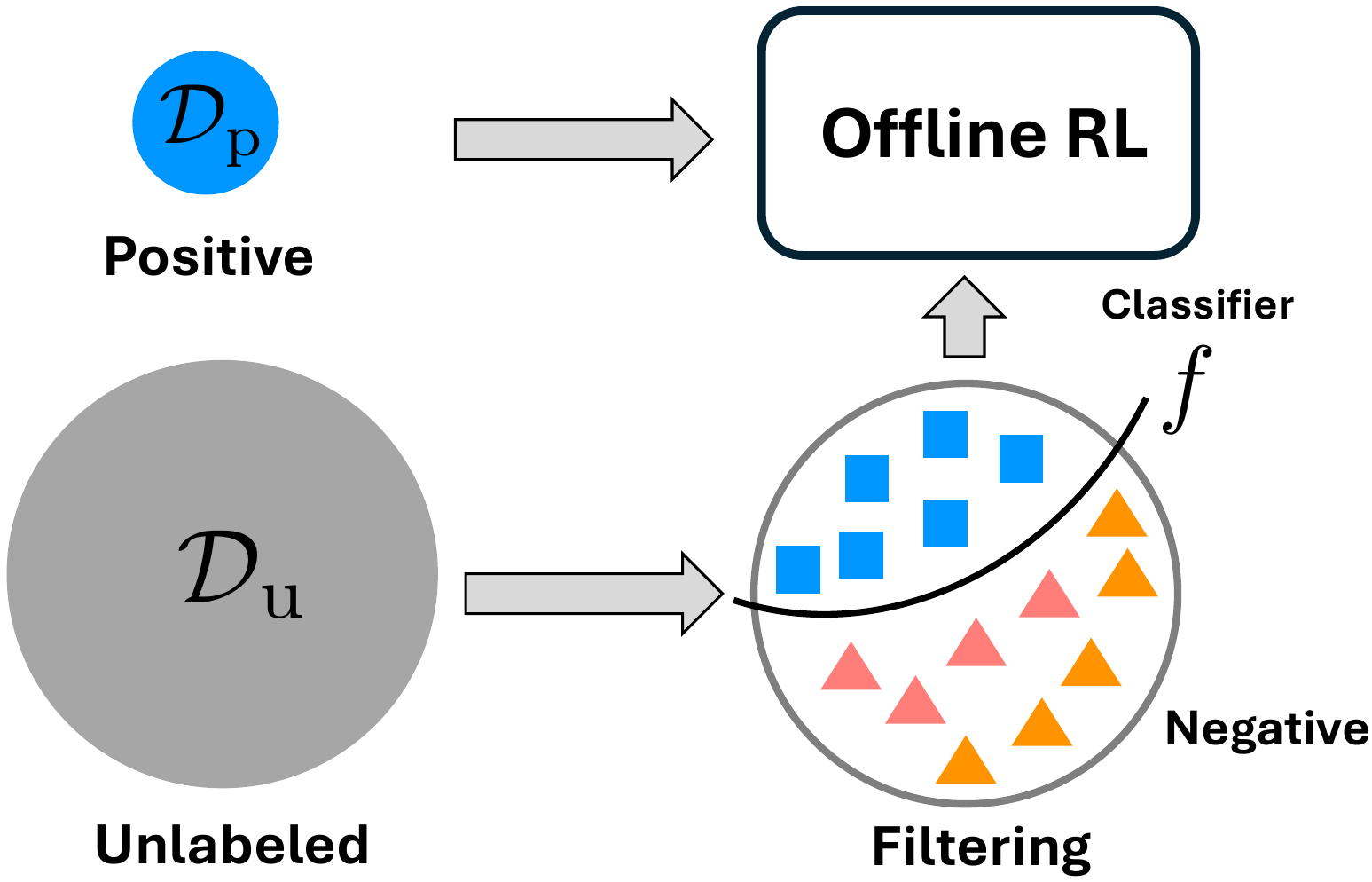}
    \caption{Diagram of our method.
 We first train a classifier $f$ using \textbf{PU learning} to distinguish positive domain data from negative domain data.
 Then, we filter the positive domain data from domain-unlabeled data by applying classifier $f$ to the domain-unlabeled dataset.
 Finally, we train a policy using off-the-shelf offline RL methods with the augmented dataset.
 }
    \label{fig:method}
\end{figure}

\subsection{Proposed Method}

The key idea of our method is to filter positive-domain data from unlabeled data by training a domain classifier that leverages the differences in transition dynamics.
Specifically, we propose a two-staged offline RL algorithm as in Figure~\ref{fig:method}.
\paragraph{Stage 1: Train a domain classifier by PU learning.}
We consider a binary classification problem where \(\cS \times \cA \times \cS'\) serves as the input space ($\cX$ in Sec.~\ref{sec:preliminaries}). The label is defined as \(Y = +1\) for the positive domain and \(Y = -1\) for the negative domains.
Since positive and negative domains differ in how they transition from \((s, a)\) to \(s'\), the tuple \((s, a,s')\) naturally captures these discrepancies, making it an effective signal for classification.
Using positive data \(\cD_\mathrm{p}\) and unlabeled data \(\cD_\mathrm{u}\), we train a classifier \(f : \cS \times \cA \times \cS' \to \{+1, -1\}\) by PU learning \citep{kiryoPositiveUnlabeledLearningNonNegative2017, sugiyama2022machine,du2015convex}.
Because \(\alpha_\mathrm{p}\) is unknown in PUORL, we estimate it using mixture proportion estimation (MPE) \citep{gargMixtureProportionEstimation2021, sugiyama2022machine}.

\paragraph{Stage 2: Data filtering and offline RL.}
We first filter the positive domain data from unlabeled data by applying classifier $f$ to the unlabeled dataset to identify instances predicted as positive, denoted by $\cD^f_{\mathrm{p}}:= \{(s, a, r, s') \in \cD_\mathrm{u}: f(s, a, s') = +1 \}$, combining it with the positive data as  $\Tilde{\cD}_\mathrm{p}:= \cD_{\mathrm{p}} \cup \cD^f_{\mathrm{p}}$.
Then, we train the policy using off-the-shelf offline RL methods with $\Tilde{\cD}_\mathrm{p}$.

The methodology details are outlined in Algo.~\ref{algo: data filtering}.

\label{sec: proposed method}
\begin{algorithm}
    \caption{Data filtering for the positive domain}\label{algo: data filtering}
    \begin{algorithmic}[1]
    \State Initialize classifier parameters $\psi$ of classifier $f$
    \State Initialize policy parameters $\theta$ and value function parameters $\phi$
    \State Initialize experience replay buffer $\cD_{\mathrm{p}}$ and  $\cD_{\mathrm{u}}$
    \State Specify epochs $K_\mathrm{PU}$, $K_\mathrm{RL}$
    \For{iteration $k \in [0, \dots, K_\mathrm{PU}]$} \Comment{PU learning routine}
    \State Update $\psi$ on $\cD_{\mathrm{p}}$ and  $\cD_{\mathrm{u}}$ by PU learning with MPE
    \EndFor
    \State $\Tilde{\cD}_{\mathrm{p}} \leftarrow \cD_{\mathrm{p}} \cup \{(s, a, r, s') \in \cD_\mathrm{u}: f_\psi(s, a, s') = +1 \}$ \Comment{Data filtering}
    \For{iteration $k \in [0, \dots, K_\mathrm{RL}]$} \Comment{Offline RL routine}
    \State Update $\theta$ and $\phi$ on $\Tilde{\cD}_{\mathrm{p}}$ by Offline RL method
    \EndFor
    \State Output $\theta$ and $\phi$
    \end{algorithmic}
\end{algorithm}
This algorithm exhibits considerable generality, accommodating a wide range of PU learning methodologies \citep{kiryoPositiveUnlabeledLearningNonNegative2017, gargMixtureProportionEstimation2021} and offline RL algorithm~\citep{kumarconservative2021, kostrikovOfflineReinforcementLearning2021, fujimoto2021a, fujimoto2023sale,tarasov2023revisiting}, allowing practitioners to choose the most suitable methods for their specific problem.
An accurate classifier is necessary for the data filtering to work well.
Conversely, less accurate classifiers result in the inclusion of negative-domain data in the filtered data $\cD^f_{\mathrm{p}}$, potentially leading to a performance decline due to the different dynamics.
\section{Experiment}
\label{sec: experiment}
We conduct experiments under various settings to investigate the following four questions:
(i) Can the PU learning method accurately classify the domain from PU-formatted data?
(ii) Can our method improve performance by augmenting positive data in various domain shift settings?
(iii) How does the magnitude of the dynamics shift affect performance?
(iv) How does the different quality of the negative-domain data affect the performance?
We first explain the setup of our experiments and, subsequently, report the results.
\subsection{Experimental Setup}
\paragraph{Dataset.}
We utilized the modified version of D4RL benchmark~\citep{fu2021d4rl} with dynamics shift, focusing on three control tasks: Halfcheetah, Hopper, and Walker2d. D4RL provides four different data qualities for each task: medium-expert (ME), medium-replay (MR), medium (M), and random (R).
To examine the impact of dynamics shift on performance, we considered three types of dynamics shifts between positive and negative domains: \textbf{body mass shift}, \textbf{mixture shift}, and \textbf{entire body shift}.
In all scenarios, we set the total number of samples to 1 million and maintained a 3:7 positive-to-negative ratio.
We explored two labeled ratios: \textbf{0.03} and \textbf{0.01}, where only 30K and 10K samples were labeled positive, respectively.
In the main text, we report the results with the labeled ratio of 0.01 and put the results with the labeled ratio of 0.03 in App.~\ref{app: supplimental_result}.

We used the dataset provided by \cite{liu2022dara} for the body mass shift and mixture shift.
In body mass shift, the mass of specific body parts in the negative domain was modified.
For the mixture shift, we mixed the data with body mass shift and data with joint noise with equal proportions to test whether our method can handle multiple negative domains.
We prepared the entire body shift with Halfcheetah and Walker2d to test the performance with a large dynamics shift.
Halfcheetah and Walker2d were paired as positive and negative domains in the entire body shift due to their entirely different body structures, yet they have the same state space of 17 dimensions.

To explore the effect of data quality on performance, we examined various combinations of data qualities, using abbreviations separated by a slash to denote pairs of positive and negative data with varying qualities, e.g., ME/ME, for medium-expert quality in both domains.

\paragraph{Offline RL algorithms and PU learning methods.}
We selected TD3+BC \citep{fujimoto2021a} and IQL \citep{kostrikovOfflineReinforcementLearning2021} as our offline RL methods due to their widespread use and computational efficiency.
We used the implementation of TD3+BC and IQL from JAX-CORL \citep{nishimori2024jaxcorl} and used the default hyperparameters for all experiments.
The main results presented below pertain to TD3+BC. The results for IQL are reported in App.~\ref{app: supplimental_result: iql}.
We trained the agent for 1 million steps and reported the average and 95\% confidence interval of averaged evaluation results over 10 episodes and 10 different seeds for each setting.

For PU learning, $\mathrm{TED}^\mathrm{n}$ \citep{gargMixtureProportionEstimation2021} was chosen owing to its effectiveness with neural networks (App.~\ref{app: explain on PU}) and used the official implementation provided by the authors.
We trained the classifier for 100 epochs and reported the average and standard deviation of the test accuracy over 5 seeds.
For more details, refer to App.~\ref{app: experimental_setup}.

\paragraph{Baselines.}
To evaluate our method's efficacy, we established five baselines for comparison: \textit{Only-Labeled-Positive (OLP)}, \textit{Sharing-All}, \textit{Dynamic-Aware Reward Augmentation (DARA)} \citep{liu2022dara}, \textit{Info-Gap Data Filtering (IGDF)} \citep{wen2024contrastive} and \textit{Oracle}.
The OLP baseline, utilizing only labeled positive data (only 1\%--3\% of the entire dataset), avoided dynamics shifts' issues at the expense of using a significantly reduced dataset size.
This comparison assessed the benefit of augmenting data volume through our filtering method.
The Sharing-All baseline employed positive and unlabeled data without preprocessing for offline RL, offering broader data coverage but posing the risk of performance degradation due to dynamics shifts. This comparison aimed to explore the impact of dynamics shifts and how our filtering technique can mitigate these effects.
The Oracle baseline, training policy with positively labeled data, and all positive data within the unlabeled data provide the ideal performance our method strives to achieve.

In addition to those naive baselines, we also compared our method with cross-domain adaptation methods designed to improve performance in the target domain by leveraging source domain data with different dynamics.
For these methods, we used the positive data as the target data and the unlabeled data as the source domain data.
We chose two methods, DARA and IGDF, which apply to any offline RL algorithms and are, thereby, good candidates for comparison with our plug-and-play method.
This comparison aimed to examine whether PUORL, where we have domain-unlabeled data alongside a limited amount of labeled target data, negatively impacts the performance of cross-domain adaptation methods.
If such a decline occurs, it highlights the need for specialized methods, such as PU-based filtering, to handle this scenario effectively.
For both algorithms, we re-implemented the algorithm in JAX \citep{jax2018github} for parallelized training referring to the official implementations. For more details, refer to App.~\ref{app: da baseline}.

\begin{table}[t] 
\centering
\caption{
 The results of the PU classifier in the body mass shift with labeled ratio = $0.01$ and $0.03$.
 For each setting, we reported the average and standard deviation of the test accuracy over 5 seeds.
}
\small
\scalebox{0.91}{
\begin{tabular}{l l c c c c}
\hline
\textbf{Env} & Ratio & ME/ME & ME/R & M/M & M/R \\
\hline
\multirow{2}{*}{\textbf{Hopper}} & $0.01$ & $99.54\pm 0.06$ & $99.23 \pm 0.08$ & $99.77\pm0.14$ & $99.33\pm0.07$ \\
& $0.03$ & $99.72 \pm 0.06$ & $99.89 \pm 0.03$  & $99.90 \pm 0.03$ & $99.32 \pm 0.05$  \\
\hline
\multirow{2}{*}{\textbf{Halfcheetah}} & $0.01$ & $99.48\pm0.04$ & $99.45 \pm 0.11$ & $99.38\pm0.18$ & $99.33\pm0.06$ \\
& $0.03$ & $99.63 \pm 0.03$ & $99.70 \pm 0.10$ & $99.66 \pm 0.06$ & $99.43 \pm 0.07$\\
\hline
\multirow{1}{*}{\textbf{Walker2d}} & $0.01$ & $99.00\pm0.03$ & $98.43 \pm 0.04$ & $98.36\pm0.02$ & $99.69\pm0.10$ \\
& $0.03$ & $99.64 \pm 0.02$ & $99.49 \pm 0.11$ & $98.41 \pm 0.06$ & $99.39 \pm 0.08$\\
\hline
\end{tabular}
}
\label{table:body_mass_shift_acc_df}
\end{table}

\begin{table}[t]
\centering
\caption{The average normalized score and 95\% confidence interval calculated by the results from 10 different seeds in body mass shift (labeled ratio = $0.01$) with TD3+BC. Of feasible methods (OLP, Sharing-All, DARA, IGDF, Ours), the best average is in \textbf{\textcolor{blue}{blue}}. Separated by the double vertical line, we report Oracle as a reference.}
\scalebox{0.87}{
\begin{tabular}{l l c c c c c || c }
\hline
\multicolumn{2}{c}{\textbf{Body mass shift}} & \multicolumn{6}{c}{} \\
\hline
\textbf{Env} & \textbf{Quality}
& \textbf{OLP} & \textbf{Sharing-All} & \textbf{DARA} & \textbf{IGDF} & \textbf{Ours}
& \textbf{Oracle} \\
\hline

\multirow{4}{*}{\textbf{Hopper}}
& ME/ME
    & $28.6 \pm 7.1$
    & $45.7 \pm 13.0$
    & $55.5 \pm 11.9$
    & $50.4 \pm 12.8$
    & \textcolor{blue}{$\mathbf{98.3 \pm 5.9}$}
    & $98.2 \pm 8.4$
\\
& ME/R
    & $36.5 \pm 7.5$
    & $73.9 \pm 12.7$
    & $51.0 \pm 9.1$
    & $40.3 \pm 8.2$
    & \textcolor{blue}{$\mathbf{100.8 \pm 6.4}$}
    & $98.2 \pm 8.4$
\\
& M/M
    & $37.9 \pm 7.3$
    & $47.4 \pm 3.4$
    & \textcolor{blue}{$\mathbf{56.6 \pm 4.6}$}
    & $52.9 \pm 2.4$
    & $48.3 \pm 1.4$
    & $48.9 \pm 2.8$
\\
& M/R
    & $43.3 \pm 4.6$
    & $45.8 \pm 4.0$
    & $52.1 \pm 4.8$
    & $50.5 \pm 4.7$
    & \textcolor{blue}{$\mathbf{52.1 \pm 2.9}$}
    & $48.9 \pm 2.8$
\\
\hline

\multirow{4}{*}{\textbf{Halfcheetah}}
& ME/ME
    & $17.6 \pm 3.1$
    & \textcolor{blue}{$\mathbf{80.8 \pm 2.1}$}
    & $27.2 \pm 3.1$
    & $21.3 \pm 5.0$
    & $75.3 \pm 10.2$
    & $86.9 \pm 4.4$
\\
& ME/R
    & $17.0 \pm 2.7$
    & $72.5 \pm 4.4$
    & $3.9 \pm 2.7$
    & $7.4 \pm 2.8$
    & \textcolor{blue}{$\mathbf{80.4 \pm 8.7}$}
    & $86.9 \pm 4.4$
\\
& M/M
    & $32.0 \pm 2.7$
    & $42.1 \pm 1.3$
    & $41.3 \pm 1.0$
    & $42.3 \pm 0.9$
    & \textcolor{blue}{$\mathbf{48.5 \pm 0.2}$}
    & $48.8 \pm 0.3$
\\
& M/R
    & $32.3 \pm 3.0$
    & $37.8 \pm 10.2$
    & $11.3 \pm 5.3$
    & $8.6 \pm 3.7$
    & \textcolor{blue}{$\mathbf{48.9 \pm 0.2}$}
    & $48.8 \pm 0.3$
\\
\hline

\multirow{4}{*}{\textbf{Walker2d}}
& ME/ME
    & $9.3 \pm 4.4$
    & $88.5 \pm 0.6$
    & $37.1 \pm 14.8$
    & $59.6 \pm 17.3$
    & \textcolor{blue}{$\mathbf{108.2 \pm 0.4}$}
    & $108.5 \pm 0.4$
\\
& ME/R
    & $15.9 \pm 5.8$
    & $78.0 \pm 24.1$
    & $2.6 \pm 1.8$
    & $4.5 \pm 2.2$
    & \textcolor{blue}{$\mathbf{108.1 \pm 0.8}$}
    & $108.5 \pm 0.4$
\\
& M/M
    & $16.4 \pm 7.0$
    & $81.2 \pm 0.8$
    & $37.0 \pm 11.3$
    & $41.7 \pm 7.6$
    & \textcolor{blue}{$\mathbf{83.2 \pm 2.2}$}
    & $84.6 \pm 0.6$
\\
& M/R
    & $21.3 \pm 7.9$
    & $80.0 \pm 2.1$
    & $1.2 \pm 1.1$
    & $0.9 \pm 1.4$
    & \textcolor{blue}{$\mathbf{84.0 \pm 0.3}$}
    & $84.6 \pm 0.6$
\\
\hline

\end{tabular}
}
\label{table:body_mass_shift_rl}
\end{table}

\begin{table}[t]
\centering
\caption{The average normalized score and 95\% confidence interval from 10 seeds in mixture shift (labeled ratio = $0.01$) with TD3+BC. The format is the same as the table for body mass shift.}
\scalebox{0.87}{
\begin{tabular}{l l c c c c c || c }
\hline
\multicolumn{2}{c}{\textbf{Mixture shift}} & \multicolumn{6}{c}{} \\
\hline
\textbf{Env} & \textbf{Quality}
& \textbf{OLP} & \textbf{Sharing-All} & \textbf{DARA} & \textbf{IGDF} & \textbf{Ours}
& \textbf{Oracle} \\
\hline

\multirow{4}{*}{\textbf{Hopper}}
& ME/ME
    & $26.8 \pm 6.2$
    & $73.0 \pm 18.6$
    & $53.4 \pm 8.8$
    & $42.4 \pm 8.9$
    & \textcolor{blue}{$\mathbf{92.6 \pm 9.7}$}
    & $96.4 \pm 8.2$
\\
& ME/R
    & $24.3 \pm 7.0$
    & $84.9 \pm 15.8$
    & $43.6 \pm 7.6$
    & $42.5 \pm 10.9$
    & \textcolor{blue}{$\mathbf{97.0 \pm 7.5}$}
    & $96.4 \pm 8.2$
\\
& M/M
    & $40.6 \pm 3.1$
    & \textcolor{blue}{$\mathbf{56.8 \pm 7.9}$}
    & $55.4 \pm 4.6$
    & $55.4 \pm 5.8$
    & $46.9 \pm 1.6$
    & $45.9 \pm 1.5$
\\
& M/R
    & $42.8 \pm 2.2$
    & $43.7 \pm 2.9$
    & $44.9 \pm 4.6$
    & \textcolor{blue}{$\mathbf{49.3 \pm 2.8}$}
    & $48.7 \pm 1.5$
    & $45.9 \pm 1.5$
\\
\hline

\multirow{4}{*}{\textbf{Halfcheetah}}
& ME/ME
    & $19.5 \pm 5.2$
    & $78.6 \pm 2.1$
    & $28.6 \pm 3.6$
    & $29.9 \pm 3.7$
    & \textcolor{blue}{$\mathbf{82.4 \pm 6.8}$}
    & $81.3 \pm 9.6$
\\
& ME/R
    & $19.0 \pm 3.1$
    & \textcolor{blue}{$\mathbf{82.0 \pm 4.8}$}
    & $11.1 \pm 4.0$
    & $9.3 \pm 1.9$
    & $78.6 \pm 8.5$
    & $81.3 \pm 9.6$
\\
& M/M
    & $35.8 \pm 2.1$
    & $48.1 \pm 1.3$
    & $39.8 \pm 2.5$
    & $40.4 \pm 2.9$
    & \textcolor{blue}{$\mathbf{48.7 \pm 0.2}$}
    & $48.7 \pm 0.2$
\\
& M/R
    & $32.5 \pm 2.0$
    & \textcolor{blue}{$\mathbf{51.7 \pm 1.4}$}
    & $14.7 \pm 3.2$
    & $16.8 \pm 4.4$
    & $48.8 \pm 0.3$
    & $48.7 \pm 0.2$
\\
\hline

\multirow{4}{*}{\textbf{Walker2d}}
& ME/ME
    & $7.0 \pm 3.1$
    & $104.4 \pm 3.5$
    & $49.1 \pm 20.2$
    & $46.0 \pm 11.9$
    & \textcolor{blue}{$\mathbf{107.6 \pm 2.0}$}
    & $108.5 \pm 0.4$
\\
& ME/R
    & $16.3 \pm 6.3$
    & $107.2 \pm 18.5$
    & $25.2 \pm 4.9$
    & $37.6 \pm 6.5$
    & \textcolor{blue}{$\mathbf{108.7 \pm 0.3}$}
    & $108.5 \pm 0.4$
\\
& M/M
    & $17.3 \pm 7.2$
    & $79.8 \pm 1.6$
    & $55.1 \pm 13.5$
    & $56.4 \pm 12.4$
    & \textcolor{blue}{$\mathbf{84.3 \pm 1.5}$}
    & $84.8 \pm 1.4$
\\
& M/R
    & $19.1 \pm 7.3$
    & $78.7 \pm 2.1$
    & $29.6 \pm 11.8$
    & $41.6 \pm 6.8$
    & \textcolor{blue}{$\mathbf{83.0 \pm 3.5}$}
    & $84.8 \pm 1.4$
\\
\hline

\end{tabular}
}
\label{table:mixture_shift_rl}
\end{table}

\begin{table}[t]
\centering
\caption{The average normalized score and 95\% confidence interval from 10 seeds in entire body shift (labeled ratio = $0.01$) with TD3+BC. The format is the same as the table for body mass shift.}
\scalebox{0.87}{
\begin{tabular}{l l c c c c c || c }
\hline
\multicolumn{2}{c}{\textbf{Entire body shift}} & \multicolumn{6}{c}{} \\
\hline
\textbf{Env} & \textbf{Quality}
& \textbf{OLP} & \textbf{Sharing-All} & \textbf{DARA} & \textbf{IGDF} & \textbf{Ours}
& \textbf{Oracle} \\
\hline

\multirow{2}{*}{\textbf{Halfcheetah}}
& ME/ME
    & $18.4 \pm 3.0$
    & $54.0 \pm 4.8$
    & $14.6 \pm 4.7$
    & $15.2 \pm 5.0$
    & \textcolor{blue}{$\mathbf{80.2 \pm 10.6}$}
    & $84.7 \pm 4.9$
\\
& ME/R
    & $21.3 \pm 2.6$
    & $33.8 \pm 10.2$
    & $9.9 \pm 2.1$
    & $16.5 \pm 4.0$
    & \textcolor{blue}{$\mathbf{89.1 \pm 4.2}$}
    & $84.7 \pm 4.9$
\\

\hline
\end{tabular}
}
\label{table:entire_body_shift_rl}
\end{table}

\subsection{Results}
\label{sec:results}

We now present the experimental findings, organized around the four key questions posed in
Section~\ref{sec: experiment}. Unless stated otherwise, all offline RL experiments use TD3+BC with
a labeled ratio of 0.01. Full results for additional settings and labeled ratios are provided in
Appendix~\ref{app: supplimental_result}.

\paragraph{(i) PU classification performance.}
\label{sec:pu_classify}
Table~\ref{table:body_mass_shift_acc_df} reports the test accuracy of
our PU classifier (based on $\mathrm{TED}^\mathrm{n}$; \citep{gargMixtureProportionEstimation2021})
for Hopper, Halfcheetah, and Walker2d under body mass shift. The accuracy exceeds 98\% in all cases,
indicating that the classifier accurately distinguishes positive-domain data from unlabeled data.
Similar performance appears under mixture shift and entire body shift, as detailed in
Appendix~\ref{app: supplimental_result: classifier}. These findings suggest that the data
filtering employed by our method is highly reliable across various shift settings.

\paragraph{(ii) Policy performance with augmented positive data.}
\label{sec:policy_perf}
Tables~\ref{table:body_mass_shift_rl}--\ref{table:entire_body_shift_rl} summarize the performance of all methods under body mass shift,
mixture shift, and entire body shift. In nearly all settings, our method achieves the
highest or near-highest average normalized score among the feasible baselines (\textit{OLP, Sharing-All,
DARA, IGDF, Ours}), often approaching the performance of the \textit{Oracle} (which has access to all
positive samples).
These results confirm that our method is effective even when only a tiny fraction of labeled positive samples are available.

\paragraph{(iii) Effect of dynamics shift magnitude.}
\label{sec:shift_magnitude} We examine performance across body mass shift, mixture shift, and entire body shift to analyze how outcomes change with increasing domain mismatch:
\begin{itemize}
    \item \textbf{Robustness of our method.} Our method's performance remains consistently strong, showing minimal degradation under larger shifts (e.g., entire body shift in Table~\ref{table:entire_body_shift_rl}).
    \item \textbf{Sharing-All vs.\ large shift.} For smaller shifts (body mass or mixture shift), \textit{Sharing-All} can occasionally yield competitive or high scores by exploiting the broader coverage.
 However, performance falls sharply as the shift increases (entire body shift).
    \item \textbf{Domain adaptation baselines (DARA, IGDF).} Although DARA~\citep{liu2022dara} and IGDF~\citep{wen2024contrastive} are designed to handle domain differences, both are worse than Sharing-All in most scenarios and degrade further with large shifts.
A likely cause is their reliance on submodule training (e.g., domain classifiers or encoders) with very few labeled data, which can become unreliable when unlabeled data may also contain additional positive samples (App.~\ref{app: supplimental_result: td3bc}). 
\end{itemize}
These patterns highlight that large domain shifts require careful data selection; our PU-based filtering remains effective, whereas both the naive Sharing-All and the domain adaptation baselines experience performance drops due to the dynamics shift.

\paragraph{(iv) Influence of negative-domain data quality.}
\label{sec:neg_data_quality}
We analyze the influence of negative-domain data quality on the performance of our method and the baselines by comparing results with different negative-domain data quality.
For example, compare ME/ME vs.\ ME/R or M/M vs.M/R with the same positive dataset quality.
We observe:
\begin{itemize}
\item \textbf{Our method} remains robust regardless of negative-domain quality. The PU filtering
consistently prevents the inclusion of harmful transitions, resulting in stable performance gains.
\item \textbf{Sharing-All and domain adaptation baselines} degrade more significantly when the
negative-domain quality is poor (e.g., R), suggesting that merging or adapting from such data
can damage performance unless the shift and data mismatch is mild.
\end{itemize}

These findings indicate that negative-domain data quality is a key factor in the methods used to share unlabeled data. By contrast, PU-based filtering appears less sensitive to
variations in the quality.
\section{Conclusion and Future Work}
This study introduced a novel offline RL setting, positive-unlabeled offline RL (PUORL), incorporating domain-unlabeled data.
We then proposed a plug-and-play algorithmic framework for PUORL that uses PU learning to augment the positively labeled data with additional positive-domain samples from the unlabeled data.
Experiments on the D4RL benchmark showed that our approach leverages large amounts of unlabeled data to train policies, achieving strong performance.
Our method primarily focused on filtering positive data from unlabeled data and training a policy solely with the filtered samples, leaving efficient cross-domain sample sharing as a future direction.
Since PU learning is a type of weakly supervised learning (WSL), we believe that extending this setting to other WSL problems could broaden offline RL's practical applications.

\bibliography{main}
\bibliographystyle{rlj}

\appendix

\section{Related Work}\label{app: related_work}
In this section, we provide a comprehensive overview of the related work.
\paragraph{Domain-adaptation in online RL.}
Various approaches have been proposed to tackle domain adaptation in online reinforcement learning, each leveraging different techniques to handle variations in environment dynamics.
Imitation learning strategies~\citep{kim2020domain, hejna2020hierarchically} utilize expert demonstrations or hierarchical policies to guide the agent in the target domain, while domain randomization methods~\citep{slaoui2019robust, mehta2020active} train agents across diverse simulated environments to build robustness against variations.
Representation learning~\citep{xing2021domain} extracts domain-invariant features to facilitate transfer, and system identification approaches~\citep{clavera2018learning} learn latent parameters of the environment’s dynamics to adapt policies online.
Data-filtering techniques~\citep{xu2023cross} selectively incorporate experience from different domains to reduce negative transfer, and reward modification based on learned classifiers~\citep{eysenbachOffDynamicsReinforcementLearning2020} helps align rewards when discrepancies in dynamics lead to misaligned feedback signals.

\paragraph{RL with multiple MDPs.}
Contextual MDPs (CMDPs) formalize the RL problem with multiple environments as MDPs controlled by a variable known as a ``context'' \citep{hallak2015contextual}. Different contexts define different types of problems \citep{Kirk_2023}.
We focus on the case where the context is a binary task ID determining the dynamics.
Thanks to its generality, the CMDP can encapsulate a wide range of RL problems, such as multi-task RL \citep{zhang2020multi, li2020multi, sodhani2021multitask} and meta-RL \citep{zintgrafVariBADVeryGood2020, dorfman2021offline}.
Depending on the observability of the context, the solution to the RL problem within CMDPs differs.
We can utilize the information in policy training if the context is observable. For example, acquiring a representation of the environment using self-supervised learning \citep{sodhani2021multitask, humplik2019meta, achiam2018variational, li2020multi} is common in addressing this objective. In offline RL, MBML (Multi-task Batch RL with Metric Learning) employed metric learning to acquire a robust representation of discrete contexts in an offline setting \citep{li2020multi}.
Unlike these approaches, our method considers settings where only a subset of the data has observable contexts.

CMDPs with unobservable contexts are also known as Hidden-Parameter (HiP)-MDPs \citep{doshi-velezHiddenParameterMarkov2016, perez2020generalized}.
In HiP-MDPs, previous works typically focused on training an inference model for the context from histories of multiple time steps \citep{rakelly2019efficient, zintgrafVariBADVeryGood2020, yooSkillsRegularizedTask2022, dorfman2021offline,ackermann2024offline}. Since we consider transition-based datasets without trajectory information, such methods are not applicable in our setting.

\paragraph{Unlabeled data in RL.}
In previous work, ``unlabeled data'' refers to two settings: reward-unlabeled data and data with the quality of the behavioral policy unknown.
In the first case, the unlabeled data consist of transitions without rewards \citep{xuPositiveUnlabeledRewardLearning2021, zolnaOfflineLearningDemonstrations2020, yuHowLeverageUnlabeled2022}.
Several studies have attempted to learn the reward function from reward-unlabeled data using the PU learning technique and then utilize this learned reward function in subsequent RL routines \citep{xuPositiveUnlabeledRewardLearning2021, zolnaOfflineLearningDemonstrations2020}. In the offline multi-task RL literature, \cite{yuHowLeverageUnlabeled2022} explored conservatively using reward-unlabeled data, i.e., setting the reward of the unlabeled transitions to zero.
In our study, the label corresponds to a specific domain, while they regard the reward as a label.
In the second case, the unlabeled data is a mixture of transitions from policies of unknown quality.
In offline RL, previous works attempted to extract high-quality data from unlabeled data using PU learning
\citep{wangImprovingBehaviouralCloning2023, yan2023simple}. In our setting, labels correspond to specific domains, not the quality of the behavioral policy.

\section{Details of Experimental Setup} \label{app: experimental_setup}
\begin{table}[ht]
    \centering
    \scalebox{0.90}{
    \begin{tabular}{lcc}
    \hline
    \multicolumn{2}{c}{\textbf{TD3+BC}} \\
    \hline
    Critic Learning Rate & $3\times10^{-4}$ \\
    Actor Learning Rate & $3\times10^{-4}$ \\
    Discount Factor & 0.99 \\
    Target Update Rate & $5\times10^{-3}$ \\
    Policy Noise & 0.2 \\
    Policy Noise Clipping & (-0.5, 0.5) \\
    Policy Update Frequency & Variable \\
    TD3+BC Hyperparameter $\alpha$ & 2.5 \\
    Actor Hidden Dims & (256, 256) \\
    Critic Hidden Dims & (256, 256) \\
    \hline
    \multicolumn{2}{c}{\textbf{IQL}} \\
    \hline
    Critic Learning Rate & $3\times10^{-4}$ \\
    Actor Learning Rate & $3\times10^{-4}$ \\
    Discount Factor & 0.99 \\
    Expectile & 0.7 \\
    Temperature & 3.0 \\
    Target Update Rate & $5\times10^{-3}$ \\
    Actor Hidden Dims & (256, 256) \\
    Critic Hidden Dims & (256, 256) \\
    \hline
    \end{tabular}
    }
    \caption{Hyperparameters for TD3+BC and IQL.}
    \label{table:hyperparameters_vertical}
    \end{table}

\subsection{PU Learning}
\label{app: explain on PU}
\paragraph{Explanation of $\mathrm{TED}^\mathrm{n}$ \citep{gargMixtureProportionEstimation2021}.}
Here, we briefly explain the $\mathrm{TED}^\mathrm{n}$ \citep{gargMixtureProportionEstimation2021} we used in our experiments. $\mathrm{TED}^\mathrm{n}$ consists of two subroutines for the mixture proportion estimation, Best Bin Estimation (BBE), and for PU learning, Conditional Value Ignoring Risk (CVIR).
They iterate these subroutines. Given the estimated mixture proportion $\hat{\alpha}$ by BBE, CVIR first discards $\hat{\alpha}$ samples from unlabeled data based on the output probability of being positive from the current classifier $f$.
The discarded samples are seemingly positive data. The classifier is then trained using the labeled positive data and the remaining unlabeled data.
On the other hand, in BBE, we estimate the mixture proportion using the output of the classifier $f$ with the samples in the validation dataset as inputs.

\paragraph{Training and evaluation.}
The PU learning method $\mathrm{TED}^\mathrm{n}$ involved two phases: warm-up and main training. We assigned 10 epochs for the warm-up step and 100 epochs for the main training step.
We utilized a 3-layer MLP with ReLU for the classifier's network architecture.
In our method, the trained classifier was then frozen and shared across different random seeds of offline RL training with identical data generation configurations, such as the positive-to-negative and unlabeled ratios.
We reported the average and standard deviation of the test accuracy over 5 random seeds.

\subsection{Offline RL}
For offline RL, we learned a policy with 1 million update steps. For both TD3+BC \citep{fujimoto2021a} and IQL \citep{kostrikovOfflineReinforcementLearning2021} we used the same hyperparameters for all baselines and settings (Table \ref{table:hyperparameters_vertical}).
We evaluated the offline RL agent using the normalized score provided by D4RL \citep{fu2021d4rl}.
To evaluate the offline RL routine's algorithmic stability, we trained with 10 different random seeds.
For each seed, we calculated the average normalized score over 10 episodes.
We reported the overall mean and 95\% confidence interval from these averaged scores.

\subsection{Baselines}
\label{app: da baseline}
\begin{algorithm}
\caption{DARA}\label{algo: DARA}
\begin{algorithmic}[1]
\Require Target offline data $\cD_\mathrm{t}$ and source offline data $\cD_\mathrm{s}$ and $\eta$.
\State Learn classifier $q_{\mathrm{sas}}: \cS \times \cA \times \cS \rightarrow [0, 1]$ and $q_{\mathrm{sa}}: \cS \times \cA \rightarrow [0, 1]$ from $\cD_\mathrm{t}$ and $\cD_\mathrm{s}$.
\State For all $(s,a, r, s')$ in $\cD_\mathrm{s}$:
\begin{equation}
    \Delta r(s, a, s') = \log \frac{q_\mathrm{sas}(\mathrm{source}|s, a, s')}{q_\mathrm{sas}(\mathrm{target}|s, a, s')} + \log \frac{q_\mathrm{sa}(\mathrm{source}|s, a)}{q_\mathrm{sa}(\mathrm{target}|s, a)}
\end{equation}
\begin{equation}
    r \leftarrow r - \eta \Delta r
\end{equation}
\State Learn policy with $\cD_\mathrm{t} \cup \cD_\mathrm{s}$.
\end{algorithmic}
\end{algorithm}

\begin{algorithm}
    \caption{IGDF: Info-Gap Data Filtering Algorithm}
    \label{algo:IGDF}
    \begin{algorithmic}[1]
    \Require Source offline data $\cD_\mathrm{s}$, target offline data $\cD_\mathrm{t}$, \\
    \quad Initialize policy $\pi$, value function $Q$, encoders $\phi(s, a), \psi(s')$, \\
    \quad data filter ratio $\xi$, importance ratio $\alpha$, batch size $B$.
    \vspace{0.5em}
    \State \textbf{// Contrastive Representation Learning}
    \State Optimize the contrastive objective in Eq.\,(6) to train the encoder networks $\phi(s, a)$ and $\psi(s')$.
    \vspace{0.5em}
    \State \textbf{// Data Filtering algorithm}
    \For{each gradient step}
        \State Sample a batch $b_\mathrm{src} := \{(s, a, r, s')\}^{\frac{B}{2}\xi}$ from $\cD_\mathrm{s}$
        \State Sample a batch $b_\mathrm{tar} := \{(s, a, r, s')\}^{\frac{B}{2}}$ from $\cD_\mathrm{t}$
        \State Select the top-$\xi$ samples from $b_\mathrm{src}$ ranked by $h(s, a, s') = = \exp(\phi(s, a)^\top \psi(s'))$
        \State Combine the top-$\xi$ samples from $b_\mathrm{src}$ with all samples from $b_\mathrm{tar}$
        \State Optimize the value function $Q_{\theta}$ via Eq.\,(8)
        \State Learn the policy $\pi(a \mid s)$ via offline RL algorithms
    \EndFor
    \end{algorithmic}
    \end{algorithm}

Here, we provide a detailed explanation of the Domain-Adaptation baselines. 

\paragraph{DARA.}
Here, we explain the Domain-Adaptation (DA) baseline used in Section \ref{sec: experiment}. For domain adaptation in offline RL, we utilized the Dynamics-Aware Reward Augmentation (DARA) \citep{liu2022dara}.
In domain adaptation in offline RL, we focus on the performance in a target domain $\cM_\mathrm{t}$ with a limited amount of target domain data $\cD_\mathrm{t}$. To address this scarcity, domain adaptation uses data $\cD_s$ from the source domain $\cM_\mathrm{s}$.
DARA modifies the source domain data's reward using a trained domain classifier and then utilizes this data with the modified reward for offline RL.
Lacking full domain labels in PUORL, we treated the positive data $\cD_\mathrm{p}$ as target domain data and the domain-unlabeled data $\cD_\mathrm{u}$ as source domain data, training the classifier with 5000 steps with batch size 256. We set $\eta = 0.1$ following original paper \citep{liu2022dara}.

\paragraph{IGDF.}
IGDF \citep{liu2022dara} is a method that uses the information of the source domain to improve the performance of the target domain.
IGDF filters the source domain data using encoder networks trained with contrastive learning with target domain data as positive samples and source domain data as negative samples.
Similar to DARA, this method is also plug-and-play. We set the representation dimension to $64$ and trained the encoder with 7000 steps with batch size 256. The data filter ratio $\xi$ is set to $0.75$ following the original paper \citep{wen2024contrastive}.
\newpage

\section{Extention to Reward Shift}
\label{app: extension_to_reward_shift}
To extend PUORL in for reward shift, we define the positive and negative MDPs as follows: positive MDP $\cM_\mathrm{p}:= (\cS, \cA, P, \rho, r_\mathrm{p}, \gamma)$, which we target for and negative MDPs $\{\cM^{k}_\mathrm{n}:= (\cS, \cA, P, \rho, r^{k}_{\mathrm{n}}, \gamma)\}_{k=1}^{N}$, which share the same state and action spaces and dynamics.
For each MDP, there exist fixed behavioral policies: $\pi_\mathrm{p}$ for positive MDP and $\pi^{k}_{\mathrm{n}}$ for negative MDPs. They induce the stationary distributions over the state-action pair denoted as $\mu_\mathrm{p}(s, a)$ and $\mu^{k}_\mathrm{n}(s, a)$ for all $k \in \{1,\ldots,N\}$.
We define $\mu_\mathrm{n}(s, a) := \sum_{k=1}^{N} \eta_k \mu_{\mathrm{n}}^{k}(s, a)$, where $\eta_k \in [0, 1], \sum_{k=1}^{N} \eta_k = 1$ is the MDP-mixture proportion.

We are given two datasets:
\begin{itemize}
    \item \textbf{Positive data}: explicitly labeled target-domain transitions, $\cD_\mathrm{p} := \{(s_i, a_i, r_i, s'_i, +1)\}_{i=1}^{n_\mathrm{p}}$.
    These transitions are i.i.d. samples from $\mu_\mathrm{p}(s, a)$, $r_\mathrm{p}$, and $P$.

    \item \textbf{Domain-unlabeled data}: a mixture of positive and negative-domain transitions, $\cD_\mathrm{u} := \{(s_i, a_i, r_i, s'_i)\}_{i=1}^{n_\mathrm{u}}$.
    These transitions are i.i.d.~samples from $\mu_\mathrm{u}(s, a) := \alpha_\mathrm{p} \mu_\mathrm{p}(s, a) + \alpha_\mathrm{n} \mu_\mathrm{n}(s, a)$, $r_\mathrm{n}$, and $P$.
\end{itemize}

Instead of taking transition, $(s, a, s')$, we take $(s, a, r)$ to train the classifier with PU learning based on the reward shift.

\begin{algorithm}
    \caption{Data filtering for the positive domain with reward shift}\label{algo: data filtering with reward shift}
    \begin{algorithmic}[1]
    \State Initialize classifier parameters $\psi$ of classifier $f$
    \State Initialize policy parameters $\theta$ and value function parameters $\phi$
    \State Initialize experience replay buffer $\cD_{\mathrm{p}}$ and  $\cD_{\mathrm{u}}$
    \State Specify epochs $K_\mathrm{PU}$, $K_\mathrm{RL}$
    \For{iteration $k \in [0, \dots, K_\mathrm{PU}]$} \Comment{PU learning routine}
    \State Update $\psi$ on $\cD_{\mathrm{p}}$ and  $\cD_{\mathrm{u}}$ by PU learning with MPE
    \EndFor
    \State $\Tilde{\cD}_{\mathrm{p}} \leftarrow \cD_{\mathrm{p}} \cup \{(s, a, r, s') \in \cD_\mathrm{u}: f_\psi(s, a, r) = +1 \}$ \Comment{Data filtering}
    \For{iteration $k \in [0, \dots, K_\mathrm{RL}]$} \Comment{Offline RL routine}
    \State Update $\theta$ and $\phi$ on $\Tilde{\cD}_{\mathrm{p}}$ by Offline RL method
    \EndFor
    \State Output $\theta$ and $\phi$
    \end{algorithmic}
\end{algorithm}
\newpage

\section{Supplemental result} \label{app: supplimental_result}
In this section, we present the supplementary results and discussion to provide additional insights into the main findings.

\begin{table}[t]
\centering
\caption{The average normalized score and 95\% confidence interval from 10 seeds in body mass shift (labeled ratio = 0.03) with TD3+BC. Of feasible methods (OLP, Sharing-All, DARA, IGDF, Ours), the best average is in \textbf{\textcolor{blue}{blue}}. The last column (Oracle) is for reference (ratio=0.05).}
\scalebox{0.87}{
\begin{tabular}{l l c c c c c || c }
\hline
\multicolumn{2}{c}{\textbf{Body mass shift (0.03)}} & \multicolumn{6}{c}{} \\
\hline
\textbf{Env} & \textbf{Quality}
& \textbf{OLP} & \textbf{Sharing-All} & \textbf{DARA} & \textbf{IGDF} & \textbf{Ours}
& \textbf{Oracle} \\
\hline

\multirow{4}{*}{\textbf{Hopper}}
& ME/ME
    & $50.0 \pm 10.1$
    & $52.3 \pm 8.1$
    & $89.9 \pm 13.8$
    & $71.1 \pm 10.1$
    & \textcolor{blue}{$\mathbf{90.4 \pm 9.2}$}
    & $98.2 \pm 8.4$
\\
& ME/R
    & $46.6 \pm 8.9$
    & $86.8 \pm 12.6$
    & $73.8 \pm 9.6$
    & $77.2 \pm 6.9$
    & \textcolor{blue}{$\mathbf{90.0 \pm 10.9}$}
    & $98.2 \pm 8.4$
\\
& M/M
    & $57.4 \pm 13.0$
    & $48.1 \pm 3.3$
    & \textcolor{blue}{$\mathbf{59.8 \pm 3.0}$}
    & $56.3 \pm 5.1$
    & $49.7 \pm 3.5$
    & $48.9 \pm 2.8$
\\
& M/R
    & $44.7 \pm 6.0$
    & $45.4 \pm 2.1$
    & $55.0 \pm 4.7$
    & \textcolor{blue}{$\mathbf{59.3 \pm 3.8}$}
    & $47.4 \pm 2.6$
    & $48.9 \pm 2.8$
\\
\hline

\multirow{4}{*}{\textbf{Halfcheetah}}
& ME/ME
    & $24.4 \pm 1.9$
    & \textcolor{blue}{$\mathbf{78.6 \pm 3.6}$}
    & $48.3 \pm 4.2$
    & $45.5 \pm 4.7$
    & $75.6 \pm 8.3$
    & $86.9 \pm 4.4$
\\
& ME/R
    & $25.3 \pm 2.5$
    & $70.3 \pm 7.7$
    & $23.6 \pm 2.1$
    & $22.5 \pm 4.2$
    & \textcolor{blue}{$\mathbf{82.2 \pm 6.9}$}
    & $86.9 \pm 4.4$
\\
& M/M
    & $43.4 \pm 1.6$
    & $41.0 \pm 0.7$
    & $45.9 \pm 0.3$
    & $46.0 \pm 0.3$
    & \textcolor{blue}{$\mathbf{48.7 \pm 0.2}$}
    & $48.8 \pm 0.3$
\\
& M/R
    & $45.4 \pm 0.5$
    & $39.6 \pm 8.1$
    & $46.7 \pm 1.7$
    & $45.2 \pm 1.9$
    & \textcolor{blue}{$\mathbf{48.5 \pm 0.2}$}
    & $48.8 \pm 0.3$
\\
\hline

\multirow{4}{*}{\textbf{Walker2d}}
& ME/ME
    & $71.7 \pm 26.1$
    & $87.9 \pm 0.9$
    & $101.8 \pm 8.7$
    & $103.9 \pm 4.0$
    & \textcolor{blue}{$\mathbf{108.7 \pm 0.2}$}
    & $108.5 \pm 0.4$
\\
& ME/R
    & $87.0 \pm 13.5$
    & $89.2 \pm 23.7$
    & $45.2 \pm 16.6$
    & $26.5 \pm 15.8$
    & \textcolor{blue}{$\mathbf{108.8 \pm 0.3}$}
    & $108.5 \pm 0.4$
\\
& M/M
    & $57.3 \pm 11.5$
    & $80.9 \pm 0.9$
    & $67.0 \pm 8.6$
    & $61.5 \pm 12.8$
    & \textcolor{blue}{$\mathbf{83.8 \pm 1.1}$}
    & $84.6 \pm 0.6$
\\
& M/R
    & $64.3 \pm 6.0$
    & $77.2 \pm 5.2$
    & $47.6 \pm 15.8$
    & $51.5 \pm 10.6$
    & \textcolor{blue}{$\mathbf{84.5 \pm 0.6}$}
    & $84.6 \pm 0.6$
\\
\hline

\end{tabular}
}
\label{table:body_mass_shift_rl_03}
\end{table}

\begin{table}[t]
\centering
\caption{The average normalized score and 95\% confidence interval from 10 seeds in mixture shift (labeled ratio = 0.03) with TD3+BC. Of feasible methods (OLP, Sharing-All, DARA, IGDF, Ours), the best average is in \textbf{\textcolor{blue}{blue}}. The last column (Oracle) is for reference (ratio=0.05).}
\scalebox{0.87}{
\begin{tabular}{l l c c c c c || c }
\hline
\multicolumn{2}{c}{\textbf{Mixture shift (0.03)}} & \multicolumn{6}{c}{} \\
\hline
\textbf{Env} & \textbf{Quality}
& \textbf{OLP} & \textbf{Sharing-All} & \textbf{DARA} & \textbf{IGDF} & \textbf{Ours}
& \textbf{Oracle} \\
\hline

\multirow{4}{*}{\textbf{Hopper}}
& ME/ME
    & $55.9 \pm 10.4$
    & $68.2 \pm 19.1$
    & $71.7 \pm 8.7$
    & $74.3 \pm 10.1$
    & \textcolor{blue}{$\mathbf{98.1 \pm 8.5}$}
    & $96.4 \pm 8.2$
\\
& ME/R
    & $48.8 \pm 7.7$
    & $84.6 \pm 10.5$
    & $80.5 \pm 4.3$
    & $69.2 \pm 12.9$
    & \textcolor{blue}{$\mathbf{100.7 \pm 4.1}$}
    & $96.4 \pm 8.2$
\\
& M/M
    & $45.0 \pm 8.1$
    & $50.4 \pm 5.3$
    & $55.6 \pm 2.6$
    & $52.9 \pm 4.0$
    & \textcolor{blue}{$\mathbf{87.6 \pm 8.7}$}
    & $45.9 \pm 1.5$
\\
& M/R
    & $48.6 \pm 1.9$
    & $49.6 \pm 5.9$
    & $57.8 \pm 3.2$
    & $55.1 \pm 3.5$
    & \textcolor{blue}{$\mathbf{49.2 \pm 2.0}$}
    & $45.9 \pm 1.5$
\\
\hline

\multirow{4}{*}{\textbf{Halfcheetah}}
& ME/ME
    & $24.3 \pm 4.4$
    & $82.1 \pm 1.3$
    & $41.6 \pm 5.4$
    & $43.1 \pm 8.7$
    & \textcolor{blue}{$\mathbf{80.0 \pm 9.0}$}
    & $81.3 \pm 9.6$
\\
& ME/R
    & $21.6 \pm 4.9$
    & $82.1 \pm 6.8$
    & $22.6 \pm 2.8$
    & $24.2 \pm 7.1$
    & \textcolor{blue}{$\mathbf{67.1 \pm 9.3}$}
    & $81.3 \pm 9.6$
\\
& M/M
    & $35.8 \pm 2.1$
    & $48.1 \pm 1.3$
    & $39.8 \pm 2.5$
    & $40.4 \pm 2.9$
    & \textcolor{blue}{$\mathbf{48.7 \pm 0.3}$}
    & $48.7 \pm 0.2$
\\
& M/R
    & $32.5 \pm 2.0$
    & $51.7 \pm 1.4$
    & $14.7 \pm 3.2$
    & $16.8 \pm 4.4$
    & \textcolor{blue}{$\mathbf{48.8 \pm 0.2}$}
    & $48.7 \pm 0.2$
\\
\hline

\multirow{4}{*}{\textbf{Walker2d}}
& ME/ME
    & $80.3 \pm 15.1$
    & $100.2 \pm 6.7$
    & $86.5 \pm 19.7$
    & $96.4 \pm 12.7$
    & \textcolor{blue}{$\mathbf{108.3 \pm 0.2}$}
    & $108.5 \pm 0.4$
\\
& ME/R
    & $90.8 \pm 14.4$
    & $101.8 \pm 23.3$
    & $72.1 \pm 14.4$
    & $89.4 \pm 20.4$
    & \textcolor{blue}{$\mathbf{108.6 \pm 0.3}$}
    & $108.5 \pm 0.4$
\\
& M/M
    & $61.6 \pm 8.0$
    & $81.8 \pm 2.4$
    & $62.4 \pm 11.3$
    & $71.0 \pm 7.9$
    & \textcolor{blue}{$\mathbf{83.9 \pm 0.8}$}
    & $84.8 \pm 1.4$
\\
& M/R
    & $64.2 \pm 7.8$
    & $79.1 \pm 3.7$
    & $66.3 \pm 16.7$
    & $65.7 \pm 16.8$
    & \textcolor{blue}{$\mathbf{82.8 \pm 1.7}$}
    & $84.8 \pm 1.4$
\\
\hline

\end{tabular}
}
\label{table:mixture_shift_rl_03}
\end{table}

\begin{table}[t]
\centering
\caption{The average normalized score and 95\% confidence interval from 10 seeds in entire body shift (labeled ratio = 0.03) with TD3+BC. Of feasible methods (OLP, Sharing-All, DARA, IGDF, Ours), the best average is in \textbf{\textcolor{blue}{blue}}. The last column (Oracle) is for reference (ratio=0.05).}
\scalebox{0.87}{
\begin{tabular}{l l c c c c c || c }
\hline
\multicolumn{2}{c}{\textbf{Entire body shift (0.03)}} & \multicolumn{6}{c}{} \\
\hline
\textbf{Env} & \textbf{Quality}
& \textbf{OLP} & \textbf{Sharing-All} & \textbf{DARA} & \textbf{IGDF} & \textbf{Ours}
& \textbf{Oracle} \\
\hline

\multirow{2}{*}{\textbf{Halfcheetah}}
& ME/ME
    & $23.1 \pm 3.9$
    & $51.8 \pm 5.3$
    & $25.7 \pm 4.8$
    & $28.2 \pm 3.2$
    & \textcolor{blue}{$\mathbf{82.7 \pm 5.8}$}
    & $84.7 \pm 4.9$
\\
& ME/R
    & $25.6 \pm 3.7$
    & $28.1 \pm 8.1$
    & $13.6 \pm 3.1$
    & $13.4 \pm 3.2$
    & \textcolor{blue}{$\mathbf{82.1 \pm 7.2}$}
    & $84.7 \pm 4.9$
\\

\hline
\end{tabular}
}
\label{table:entire_body_shift_rl_03}
\end{table}

\subsection{Results with TD3+BC with labeled ratio = 0.03}
\label{app: supplimental_result: td3bc}
Table~\ref{table:body_mass_shift_rl_03}--~\ref{table:entire_body_shift_rl_03} show the results of TD3+BC with the labeled ratio = $0.03$.
For all the results, our method achieves the best performance in almost all the settings, indicating its efficacy in PUORL.
Another point to note is that the performance of the domain adaptation baselines is improved compared with the labeled ratio of $0.01$, indicating the severe influence of extremely limited labeled target domain data.

\begin{table}[t]
    \centering
    \caption{The average normalized score and 95\% confidence interval from 10 seeds in body mass shift (labeled ratio = $0.01$) with IQL.}
    \scalebox{0.87}{
    \begin{tabular}{l l c c c c c || c }
    \hline
    \multicolumn{2}{c}{\textbf{Body mass shift}} & \textbf{OLP} & \textbf{Sharing-All} & \textbf{DARA} & \textbf{IGDF} & \textbf{PU} & \textbf{Oracle} \\
    \hline
    \multirow{5}{*}{\textbf{Hopper}}
    & ME/ME & $23.9\pm5.9$   & $38.99\pm14.1$  & $37.44\pm7.6$  & $29.75\pm5.84$  & \textcolor{blue}{$\mathbf{39.72\pm8.64}$}  & $54.3\pm14.9$ \\
    & ME/R  & $23.35\pm4.23$ & $7.79\pm0.16$    & $7.74\pm0.25$  & $11.21\pm5.02$  & \textcolor{blue}{$\mathbf{42.04\pm9.95}$}  & $54.3\pm14.9$ \\
    & M/M   & $37.37\pm4.58$ & $37.06\pm1.28$   & $35.73\pm1.17$ & $36.28\pm7.38$  & \textcolor{blue}{$\mathbf{56.37\pm3.83}$}  & $54.2\pm3.3$ \\
    & M/R   & \textcolor{blue}{$\mathbf{33.56\pm3.47}$} & $8.04\pm0.16$    & $21.68\pm8.98$ & $12.37\pm5.4$   & $18.66\pm8.36$  & $54.3\pm14.9$ \\
    \hline
    \multirow{5}{*}{\textbf{Halfcheetah}}
    & ME/ME & $0.7\pm0.86$   & $51.53\pm3.96$   & $54.41\pm2.17$  & $51.33\pm3.18$  & \textcolor{blue}{$\mathbf{82.72\pm3.57}$}  & $87.3\pm2.7$ \\
    & ME/R  & $-0.11\pm0.47$ & $26.95\pm7.54$   & $47.71\pm7.73$ & $38.39\pm4.55$  & \textcolor{blue}{$\mathbf{84.83\pm4.06}$}  & $87.3\pm2.7$ \\
    & M/M   & $3.89\pm1.62$  & $37.3\pm0.2$              & $36.93\pm0.22$ & $36.43\pm0.7$   & \textcolor{blue}{$\mathbf{46.33\pm0.46}$}  & $46.5\pm0.1$ \\
    & M/R   & $6.31\pm3.66$  & $41.64\pm2.38$   & $43.56\pm0.5$  & $41.18\pm1.9$   & \textcolor{blue}{$\mathbf{46.57\pm0.13}$}  & $46.5\pm0.1$ \\
    \hline
    \multirow{5}{*}{\textbf{Walker2d}}
    & ME/ME & $4.3\pm2.75$   & $90.68\pm0.38$   & $88.92\pm6.82$ & $96.73\pm7.74$  & \textcolor{blue}{$\mathbf{110.32\pm0.78}$} & $109.1\pm1.4$ \\
    & ME/R  & $6.64\pm6.22$  & $65.18\pm12.23$  & $62.04\pm17.2$ & $66.61\pm10.58$ & \textcolor{blue}{$\mathbf{88.57\pm14.48}$} & $109.1\pm1.4$ \\
    & M/M   & $14.42\pm5.57$ & \textcolor{blue}{$\mathbf{82.77\pm0.45}$}   & $82.42\pm0.63$ & $74.85\pm6.35$  & $73.49\pm9.69$  & $75.6\pm5.2$ \\
    & M/R   & $4.79\pm3.4$   & $52.22\pm7.44$   & $47.96\pm6.59$ & \textcolor{blue}{$\mathbf{54.33\pm10.88}$} & $50.34\pm19.19$ & $75.6\pm5.2$ \\
    \hline
    \end{tabular}
    }
    \label{table:body_mass_shift_iql_0.01}
\end{table}

\begin{table}[t]
    \centering
    \caption{The average normalized score and 95\% confidence interval from 10 seeds in mixture shift (labeled ratio = $0.01$) with IQL.}
    \scalebox{0.87}{
    \begin{tabular}{l l c c c c c || c }
    \hline
    \multicolumn{2}{c}{\textbf{Mixture shift}} & \textbf{OLP} & \textbf{Sharing-All} & \textbf{DARA} & \textbf{IGDF} & \textbf{PU} & \textbf{Oracle} \\
    \hline
    \multirow{5}{*}{\textbf{Hopper}}
    & ME/ME & $20.43\pm6.53$ & $24.58\pm4.47$  & $43.0\pm10.54$ & \textcolor{blue}{$\mathbf{43.08\pm11.08}$} & $35.28\pm8.5$  & $54.3\pm14.9$ \\
    & ME/R  & $19.1\pm3.74$  & $22.08\pm3.58$  & \textcolor{blue}{$\mathbf{36.36\pm7.83}$}  & $28.43\pm8.29$  & $29.03\pm6.45$  & $54.3\pm14.9$ \\
    & M/M   & $29.96\pm3.98$ & \textcolor{blue}{$\mathbf{61.82\pm8.82}$}  & $54.86\pm9.64$  & $51.55\pm6.13$  & $50.1\pm3.34$   & $54.2\pm3.3$ \\
    & M/R   & $33.98\pm3.15$ & $44.78\pm7.52$  & \textcolor{blue}{$\mathbf{50.47\pm2.51}$}  & $47.2\pm2.13$   & $45.86\pm1.44$  & $54.2\pm3.3$ \\
    \hline
    \multirow{5}{*}{\textbf{Halfcheetah}}
    & ME/ME & $0.26\pm0.67$  & $62.46\pm1.43$ & $56.73\pm3.36$ & $57.48\pm4.63$ & \textcolor{blue}{$\mathbf{69.5\pm2.97}$} & $87.3\pm2.7$ \\
    & ME/R  & $1.13\pm1.09$  & $57.94\pm7.1$  & $49.84\pm9.63$ & $51.46\pm8.41$ & \textcolor{blue}{$\mathbf{72.83\pm4.74}$} & $87.3\pm2.7$ \\
    & M/M   & $5.58\pm2.32$  & $48.05\pm0.64$ & \textcolor{blue}{$\mathbf{48.43\pm0.42}$} & $46.36\pm2.09$ & $46.54\pm0.19$ & $46.5\pm0.1$ \\
    & M/R   & $7.16\pm4.39$  & $44.97\pm0.46$ & $44.84\pm1.45$ & $43.16\pm1.04$ & \textcolor{blue}{$\mathbf{46.58\pm0.21}$} & $46.5\pm0.1$ \\
    \hline
    \multirow{5}{*}{\textbf{Walker2d}}
    & ME/ME & $3.28\pm2.08$  & $93.95\pm19.68$ & $96.8\pm13.09$  & $93.13\pm11.37$ & \textcolor{blue}{$\mathbf{108.46\pm2.65}$} & $109.1\pm1.4$ \\
    & ME/R  & $6.33\pm3.13$  & $93.38\pm8.08$  & $89.98\pm12.61$ & $88.98\pm13.59$ & \textcolor{blue}{$\mathbf{98.96\pm12.36}$} & $109.1\pm1.4$ \\
    & M/M   & $3.55\pm2.43$  & $74.68\pm2.73$  & \textcolor{blue}{$\mathbf{72.85\pm5.38}$}  & $64.14\pm8.58$  & $64.45\pm13.22$ & $75.6\pm5.2$ \\
    & M/R   & $11.84\pm7.15$ & $50.24\pm6.64$  & $60.78\pm7.39$  & $59.43\pm7.88$  & \textcolor{blue}{$\mathbf{62.15\pm18.06}$} & $75.6\pm5.2$ \\
    \hline
    \end{tabular}
    }
    \label{table:mixture_shift_iql_0.01}
\end{table}

\begin{table}[t]
    \centering
    \caption{The average normalized score and 95\% confidence interval from 10 seeds in halfcheetah vs walker2d shift (labeled ratio = $0.01$) with IQL.}
    \scalebox{0.87}{
    \begin{tabular}{l l c c c c c || c }
    \hline
    \multicolumn{2}{c}{\textbf{Halfcheetah vs Walker2d}} & \textbf{OLP} & \textbf{Sharing-All} & \textbf{DARA} & \textbf{IGDF} & \textbf{PU} & \textbf{Oracle} \\
    \hline
    \multirow{3}{*}{\textbf{Halfcheetah}}
    & ME/ME & $0.28\pm0.37$ & $40.78\pm3.42$ & $52.55\pm4.95$ & $53.96\pm4.46$ & $\textcolor{blue}{\mathbf{89.31\pm1.92}}$ & $87.3\pm2.7$ \\
    & ME/R  & $0.07\pm0.57$ & $35.64\pm3.58$ & $36.93\pm3.33$ & $31.12\pm4.52$ & $\textcolor{blue}{\mathbf{86.73\pm2.91}}$ & $87.3\pm2.7$ \\
    \hline
    \end{tabular}
    }
    \label{table:halfcheetah_vs_walker2d_iql_0.01}
\end{table}

\subsection{Results with IQL}
\label{app: supplimental_result: iql}
Here, we provide the experimental results with IQL \citep{kostrikovOfflineReinforcementLearning2021}.
Table~\ref{table:body_mass_shift_iql_0.01}--\ref{table:halfcheetah_vs_walker2d_iql_0.01} show the results with the labeled ratio = $0.01$.
The results show that our method achieves the best performance in 17 out of 26 settings.
Overall, the results with hopper are unstable and worse for all methods, indicating that the performance of IQL is sensitive in Hopper with limited data (30\% in maximum).

\begin{table}[t]
    \centering
    \caption{
        The results of the PU classifier in the mixture shift with labeled ratio = $0.01$ and $0.03$.
        For each setting, we reported the average and standard deviation of the test accuracy over 5 seeds.
    }
    \small
    \scalebox{0.91}{
    \begin{tabular}{l l c c c c}
    \hline
    \textbf{Env} & Ratio & ME/ME & ME/R & M/M & M/R \\
    \hline
    \multirow{2}{*}{\textbf{Hopper}}
    & $0.01$ & $98.92\pm0.54$ & $98.91\pm0.14$ & $99.33\pm0.20$ & $99.21\pm0.08$ \\
    & $0.03$ & $99.44\pm0.11$ & $99.22\pm0.09$ & $99.79\pm0.11$ & $99.42\pm0.05$ \\
    \hline
    \multirow{2}{*}{\textbf{Halfcheetah}}
    & $0.01$ & $99.43\pm0.10$ & $99.42\pm0.10$ & $99.38\pm0.05$ & $99.35\pm0.03$ \\
    & $0.03$ & $99.63\pm0.04$ & $99.56\pm0.03$ & $99.39\pm0.02$ & $99.32\pm0.19$ \\
    \hline
    \multirow{1}{*}{\textbf{Walker2d}}
    & $0.01$ & $98.49\pm0.14$ & $98.02\pm0.16$ & $98.63\pm0.19$ & $98.05\pm0.12$ \\
    & $0.03$ & $99.00\pm0.07$ & $98.83\pm0.10$ & $99.26\pm0.08$ & $98.81\pm0.25$ \\
    \hline
    \end{tabular}
    }
    \label{table:mixture_shift_acc_df}
\end{table}

\begin{table}[t]
    \centering
    \caption{
        The results of the PU classifier in the entire body shift with labeled ratio = $0.01$ and $0.03$.
        For each setting, we reported the average and standard deviation of the test accuracy over 5 seeds.
    }
    \small
    \scalebox{0.91}{
    \begin{tabular}{l l c c c c}
    \hline
    \textbf{Env} & Ratio & ME/ME & ME/R & M/M & M/R \\
    \hline
    \multirow{2}{*}{\textbf{Halfcheetah}}
    & $0.01$ & $99.76\pm0.28$ & $99.87\pm0.15$ & $99.79\pm0.11$ & $99.74\pm0.13$ \\
    & $0.03$ & $99.98\pm0.01$ & $99.93\pm0.04$ & $99.95\pm0.06$ & $99.96\pm0.21$ \\
    \hline
    \end{tabular}
    }
    \label{table:entire_body_shift_acc_df}
\end{table}

\subsection{Classifier Performance}
\label{app: supplimental_result: classifier}
Here, we review the performance of the classifier under the mixture shift.
Seeing Table \ref{table:mixture_shift_acc_df}--\ref{table:entire_body_shift_acc_df}, we can see that the PU classifier achieved higher than 98\%  accuracy, demonstrating the efficacy of PU learning under mixture shift and entire body shift.

\end{document}